\documentclass[runningheads]{llncs}
\usepackage[T1]{fontenc}
\usepackage{amsmath}   

\usepackage{newtxtext}       %
\usepackage[varvw]{newtxmath}       

\usepackage{graphicx}
\usepackage{booktabs}  
\begin{document}
\title{Generate-then-Rank: Learning Symbolic Goal Interpretation for Embodied Agents}
\titlerunning{Generate-then-Rank for Symbolic Goal Interpretation}
%
\author{Jisen Lei \and Huizhi Liang}
\authorrunning{J. Lei and H. Liang}
%
\institute{School of Computing, Newcastle University,
Newcastle upon Tyne, United Kingdom\\
\email{leijisena@gmail.com}}
\maketitle              

\begin{abstract}
Goal interpretation is to translate a natural language instruction into a formal symbolic goal that a planner can execute. It is critical for embodied agents because any error here propagates to every downstream planning and action step. Doing it well requires both generating suitable candidates and reliably selecting among them.
We present a two-stage generate-then-rank approach to Goal Interpretation in the Embodied Agent Interface (EAI). First, supervised fine-tuning adapts a Large Language Model (LLM) to EAI's output convention. Second, Candidate-Output Ranking from Errors (CORE) learns preferences over complete outputs from the model's own generated alternatives. We conducted experiments on the household scenes of the Embodied Agent Interface Benchmark. Multiple open source LLMs including DeepSeek-V3 chat, DeepSeek-Reasoner, and Qwen2.5-7B and Qwen2.5-72B were evaluated. The proposed method outperformed the baseline models including LLMs with simple prompting, chain-of-thought, and knowledge graph based approaches.
\end{abstract}

\section{Introduction}
\label{sec:introduction}
Large language models (LLMs) increasingly connect natural-language instructions
to embodied planning and action \cite{ahn2022saycan,shridhar2021alfworld,yao2023react}.
Before planning, however, an agent must determine what terminal conditions
constitute success. Goal interpretation is to translate a natural language instruction into a formal symbolic goal that a planner can execute. It is critical for embodied agents because any error here propagates to every downstream planning and action step. The Benchmark Embodied Agent Interface (EAI) \cite{li2024embodied} formalises this goal
interpretation task as a mapping from a VirtualHome scene and instruction
to object states, directed relations, and terminal actions
\cite{li2024embodied,puig2018virtualhome}. An omitted condition, unintended
relation, or incompatible predicate argument can make downstream behaviour
incomplete or infeasible even when the planner is otherwise correct.

For example, the instruction ``wash dishes with dishwasher'' in
Fig.~\ref{fig:grace_framework} requires the dishwasher to be \texttt{CLOSED}
and \texttt{ON} and the directed relations \texttt{ON(plate, dishwasher)},
\texttt{ON(cup, dishwasher)}, and \texttt{ON(dish\_soap, dishwasher)}.
Walking to the kitchen or opening the dishwasher is an intermediate step, not
a terminal goal. A goal that omits \texttt{ON(dish\_soap, dishwasher)} is
still well formed, yet it is incomplete. A relation is also directed:
\texttt{ON(plate, dishwasher)} is correct, whereas
\texttt{ON(dishwasher, plate)} is wrong. Well-formed output alone therefore
does not ensure correct goal interpretation.

Direct generation provides a useful baseline \cite{li2024embodied}, but the
single output a model returns is not necessarily the best goal it can
generate: sampling several outputs can yield a more complete goal.
Task-specific supervised fine-tuning (SFT) can adapt an LLM to the
required output format and sparse terminal-goal convention \cite{qwen2024report,hu2021lora}.
SFT alone, however, optimises reference-token likelihood rather than an
explicit preference between complete candidate goals. Sequence-level ranking
has addressed related mismatches in text generation \cite{liu-etal-2022-brio,zhao2023slichf}.
Our candidate analysis motivates studying this mismatch for symbolic goal
interpretation: a better goal may already be generated but remain unselected.

Our approach comprises two learned stages, shown in
Fig.~\ref{fig:grace_framework}. Stage~1 is supervised fine-tuning for goal
generation: it takes the scene and instruction and produces several complete
candidate goals. Stage~2 is learning to rank candidate outputs, which we call
Candidate-Output Ranking from Errors (CORE): it takes the input and its
candidate pool, assigns every candidate a sequence score, and outputs the
highest-scoring goal. CORE is learned rather than rule-based. During training
it compares the model's own better and worse candidates and adjusts the
scorer so that higher-quality complete Node--Edge--Action outputs receive
higher scores, while a supervised anchor retains the reference target.
Its contribution is the application of error-derived candidate ranking to
symbolic goals, rather than a claim to invent supervised or preference learning.

Our main contributions are as follows:
(i) a two-stage generate-then-rank approach that combines generation and
ranking to improve goal interpretation with large language models;
(ii) a training procedure that first applies supervised fine-tuning and then
learns to rank candidates from the model's own errors with a contrastive
objective; and
(iii) experiments on the household scenes of the Embodied Agent Interface
benchmark, comparing several LLMs and baselines that include supervised
fine-tuning, candidate-level preference optimisation, and knowledge-graph-based
methods.

\section{Related Work}
\label{sec:related_work}
Embodied-language systems ground instructions in affordances, generate action
programs, and interleave reasoning with interaction
\cite{ahn2022saycan,shridhar2021alfworld,yao2023react}. EAI instead isolates
four interfaces: goal interpretation, subgoal decomposition, action sequencing,
and transition modelling \cite{li2024embodied}. Goal interpretation predicts sparse terminal
Node, Edge, and Action records rather than a trajectory. This distinction makes
plausible intermediate actions false positives and requires exact relation
direction, cardinality, and stopping.

Autoregressive likelihood scores token sequences, whereas EAI evaluates the
parsed final set. Candidate-ranking methods such as BRIO and SLiC-HF calibrate
complete outputs with preference and supervised signals
\cite{liu-etal-2022-brio,zhao2023slichf}; DPO learns preferences relative to a
frozen reference policy \cite{rafailov2023direct}. Cal-DPO additionally
calibrates implicit reward magnitudes \cite{xiao2024caldpo}; CORE instead uses
reference-derived quality gaps to scale pairwise margins. CORE applies this
sequence-level perspective to complete symbolic goals, using realised model
errors and an SFT anchor. Related planning research also shows the value of
structured environmental evidence \cite{rana2023sayplan,xie2023translating};
we therefore include knowledge-graph-based comparators in our experiments.
EAI itself evaluates LLMs by direct generation \cite{li2024embodied}; ranking
among complete candidate goals is not part of that protocol, which is the gap
our second stage addresses.

Sampling and self-consistency offer another way to use multiple model outputs
\cite{wang2022selfconsistency}; our selector instead scores deduplicated,
complete symbolic candidates. Constrained decoding, exemplified by PICARD,
checks admissibility during generation \cite{scholak2021picard}. DOMINO
aligns formal constraints with subword tokenisation \cite{beurerkellner2024domino}.
Neither Cal-DPO nor these decoding methods is reproduced here.

\section{Method}
\label{sec:methodology}
\begin{figure}[!t]
  \centering
  \includegraphics[width=0.9\textwidth]{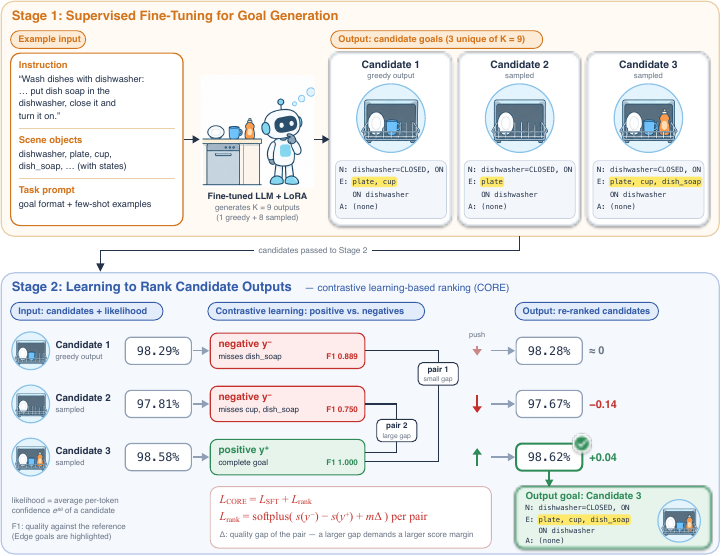}
  \Description{Two-panel diagram. Top panel, Stage 1: an example input (the
  instruction to wash dishes with a dishwasher, the scene objects, and the task
  prompt) is given to a fine-tuned LLM, drawn as a robot, which generates three
  candidate goals. Each candidate is shown as a dishwasher rack and as node, edge,
  and action goals: Candidate 1 contains a plate and a cup, Candidate 2 only a
  plate, and Candidate 3 a plate, a cup, and dish soap. Bottom panel, Stage 2: the
  candidates enter with likelihoods of 98.29, 97.81, and 98.58 percent.
  Candidates 1 and 2 are negatives and Candidate 3 is the positive in two
  contrastive pairs. After training the likelihoods are 98.28, 97.67, and 98.62
  percent, and Candidate 3, with Combined F1 1.000, is the output goal.}
  \caption{Our two-stage generate-then-rank approach on an example record.
  Stage~1: given the instruction, scene objects, and task prompt, the
  fine-tuned LLM generates $K=9$ outputs, of which three are unique candidate
  goals; each is drawn as the dishwasher contents it specifies and written as
  node (N), edge (E), and action (A) goals. Stage~2: the candidates and their
  likelihoods are the input. Contrastive learning pairs the positive, the best
  candidate, with each negative, a lower-quality candidate; the ranking loss
  raises the positive's score and lowers the negatives', more strongly for
  the pair with the larger quality gap. The output is the candidate with the
  highest likelihood after training. Likelihood is the average per-token
  probability; changes are in percentage points.}
  \label{fig:grace_framework}
\end{figure}

Figure~\ref{fig:grace_framework} gives an overview of our generate-then-rank
method: Stage~1 fine-tunes a generator that produces complete candidate goals,
and Stage~2 learns to rank those candidates and selects one. We first define the task and
then describe each stage.

\subsection{Problem Formulation}
\label{sec:problem_formulation}
For item $i$, let $x_i=\langle s_{0,i},l_{g,i}\rangle$ contain the initial
scene $s_{0,i}$ and language instruction $l_{g,i}$. Goal interpretation maps
this input to a complete symbolic goal
\begin{equation}
\label{eq:gi_mapping}
G(x_i)\rightarrow g_i,\qquad g_i=(N_i,E_i,A_i),
\end{equation}
where $N_i$ is the set of target object states, $E_i$ is the set of directed
target relations, and $A_i$ is the set of terminal task actions. These sets
specify success conditions rather than an execution trace. A model produces a
goal as a token sequence $y$; $y_i^{*}$ denotes the sequence of the reference
goal of item $i$.

For the record shown in Fig.~\ref{fig:grace_framework}, the reference goal is
\begin{equation*}
\begin{aligned}
N_i&=\{(\texttt{dishwasher},\texttt{CLOSED}),\,(\texttt{dishwasher},\texttt{ON})\},\\
E_i&=\{(\texttt{plate},\texttt{ON},\texttt{dishwasher}),\,(\texttt{cup},\texttt{ON},\texttt{dishwasher}),\\
   &\qquad(\texttt{dish\_soap},\texttt{ON},\texttt{dishwasher})\},\\
A_i&=\emptyset.
\end{aligned}
\end{equation*}
$N_i$ records an object's terminal states and $E_i$ records directed
source--relation--target triples. $A_i$ is empty because this goal requires
no action beyond what states and relations express; other goals do, for
instance a required \texttt{SLEEP} action. The goal is thus a graph with
three parts: \texttt{node goals}, \texttt{edge goals}, and \texttt{action goals}.
Components may be empty when the reference goal does not require them; an
action list is not a full plan.

\begin{table}[!t]
\centering
\caption{Notation used in the generation and ranking objectives.}
\label{tab:symbols}
\footnotesize
\setlength{\tabcolsep}{4pt}
\begin{tabular}{@{}lp{0.70\columnwidth}@{}}
\toprule
Symbol & Meaning\\
\midrule
$x_i=\langle s_{0,i},l_{g,i}\rangle$ & input of item $i$: initial scene and instruction\\
$g_i=(N_i,E_i,A_i)$ & symbolic goal with node, edge, and action goals\\
$y,\,y_t,\,y_{<t}$ & output token sequence, its token $t$, and the preceding tokens\\
$y_i^{*}$ & token sequence of the reference goal\\
$\theta,\,p_\theta$ & trainable LoRA parameters and the model's next-token distribution\\
$K,\,\mathcal C_i,\,M_i$ & number of generated outputs, candidate pool, and number of unique candidates\\
$s_\theta(y\mid x_i)$ & length-normalised sequence score of candidate $y$\\
$\hat y_i$ & selected candidate\\
$y_i^{+},\,y^{-},\,\mathcal P_i^{-}$ & positive (best) candidate, a negative (lower-quality) candidate, and the set of negatives\\
$F_{1}^{k},\,F_{1}^{\mathrm{all}}$ & F1 of component $k\in\{N,E,A\}$ and Combined F1\\
$U_i(y),\,\Delta_{i,y^{-}}$ & utility of candidate $y$ and quality gap of a pair\\
$d_{i,y^{-}},\,\delta_{i,y^{-}}$ & score difference and scaled margin of a pair\\
$\mathcal L_{\mathrm{SFT}},\,\mathcal L_{\mathrm{rank}},\,\mathcal L_{\mathrm{CORE}}$ & supervised, ranking, and combined losses\\
$\alpha,\,\gamma,\,m$ & SFT weight, ranking weight, and margin scale\\
$Q_i(y)$ & reference-based quality used by the candidate oracle\\
$\beta$ & DPO temperature\\
\bottomrule
\end{tabular}
\end{table}
\subsection{Stage 1: Supervised Fine-Tuning for Goal Generation}
\label{sec:stage1}

The generator is a pretrained LLM with LoRA adapters \cite{hu2021lora} on its attention
and MLP projections. For input $x_i$ and reference sequence
$y_i^{*}=(y^{*}_{i,1},\ldots,y^{*}_{i,|y_i^{*}|})$, prompt, system, and padding
tokens are masked and the supervised fine-tuning (SFT) objective is
\begin{equation}
\label{eq:sft_loss}
\mathcal{L}_{\mathrm{SFT},i}
=-\frac{1}{|y_i^{*}|}\sum_{t=1}^{|y_i^{*}|}\log p_\theta(y^{*}_{i,t}\mid x_i,y^{*}_{i,<t}),
\end{equation}
where the mean is over unmasked assistant tokens, $p_\theta$ is the next-token distribution of the model with trainable
parameters $\theta$, $y^{*}_{i,t}$ is the reference token at position $t$, and
$y^{*}_{i,<t}$ is its prefix. Greedy decoding provides the SFT baseline. The
supervision teaches both the output format and the EAI convention governing goal
presence, cardinality, omission, and stopping.
For a goal such as the one in Fig.~\ref{fig:grace_framework}, the reference
goal contains two node records (\texttt{name=dishwasher} with
\texttt{state=CLOSED} and \texttt{state=ON}), three edge records of the form
\texttt{from\_name=plate, relation=ON, to\_name=dishwasher}, and an empty
action list. For training records, these reference tokens form the SFT
target.
The implementation uses a causal language model and autoregressive generation:
it predicts successive tokens until a complete output is produced. It is
structured generation, not a single fixed-label classification, despite the
prompt restricting available objects and predicates.

\subsection{Stage 2: Learning to Rank Candidate Outputs}
\label{sec:stage2}

\paragraph{Why rank candidates}
Greedy decoding takes the most probable token at every step. Each choice is
locally best, but the resulting complete goal is not necessarily the best goal
the model can produce: one early token, such as the source of an Edge, fixes
the rest of the output. Sampling several complete outputs exposes
alternatives, and our candidate analysis
(Section~\ref{sec:results}) finds items whose pool contains a better goal
than the one selected. Stage~1, however, is trained only to reproduce
reference tokens; nothing teaches it to score a better complete candidate
above a worse one. Stage~2 therefore adds a learning-to-rank step. Unlike a
fixed scoring rule, its ranking function is optimised from data.

\paragraph{Input, output, and components}
As shown in the Stage~2 part of Fig.~\ref{fig:grace_framework}, the input is
$x_i$ together with its candidate pool, and the output is one selected
complete goal $\hat y_i$. The only component is a scorer: the Stage~1 model,
whose LoRA parameters continue to be trained. No separate ranking network or
classification head is added. Training changes the scorer with the CORE loss
below; inference applies the trained scorer to each candidate and returns the
highest-scoring one without reading any reference.

\paragraph{Candidate pool and sequence score}

For each input $x_i$, a fixed task-adapted generation checkpoint produces
$K=9$ outputs: one greedy completion and two groups of four sampled
completions. Exact duplicate strings are removed, yielding candidate pool
$\mathcal C_i$ with $1\leq M_i\leq K$ unique complete Node--Edge--Action goals. Deduplication occurs before scoring, so
repeated samples have no voting advantage.

Teacher forcing assigns candidate $y\in\mathcal C_i$ the length-normalised
score
\begin{equation}
\label{eq:sequence_score}
s_\theta(y\mid x_i)=\frac{1}{|y|}
\sum_{t=1}^{|y|}\log p_\theta(y_t\mid x_i,y_{<t}).
\end{equation}
The likelihood selector returns
$\hat y_i=\arg\max_{y\in\mathcal C_i}s_\theta(y\mid x_i)$. Division by token
length $|y|$ reduces automatic preference for shorter serialisations. At inference, this rule compares complete graphs rather than individual Edge
tokens. During CORE training, the same differentiable score is recomputed
under the current trainable model; its parameters are not frozen.
Length normalisation does not make scores invariant to key order, whitespace,
or other semantically equivalent serialisations. We score the archived
strings without semantic-field masking; format robustness is not established.

\paragraph{Learning from candidate errors}

CORE is a contrastive, pairwise learning-to-rank objective: it learns from
quality contrasts observed within model-generated candidate pools. For item
$i$, the positive $y_i^+$ is its highest-utility candidate and the negatives
$\mathcal P_i^-$ are the candidates with lower utility. Utility combines canonical
Combined F1 with a 0.250-weighted Edge F1 term. CORE retains an SFT anchor on
the reference output. For readability, define the score difference and positive pair margin as
\begin{equation}
\begin{aligned}
d_{i,y^-}&=s_\theta(y^-\mid x_i)-s_\theta(y_i^+\mid x_i),\\
\delta_{i,y^-}&=m\times\Delta_{i,y^-}.
\end{aligned}
\end{equation}
The ranking loss and the complete CORE loss are
\begin{equation}
\label{eq:core_loss}
\begin{aligned}
\mathcal L_{\mathrm{rank},i}
&=\frac{1}{|\mathcal P_i^-|}
\sum_{y^-\in\mathcal P_i^-}
\mathrm{softplus}(d_{i,y^-}+\delta_{i,y^-}),\\
\mathcal L_{\mathrm{CORE},i}
&=\alpha\,\mathcal L_{\mathrm{SFT},i}+\gamma\,\mathcal L_{\mathrm{rank},i},
\end{aligned}
\end{equation}
with the reference-derived quality gap
\begin{equation}
\begin{aligned}
\Delta_{i,y^-}&=\max\{0.01,\,U_i(y_i^+)-U_i(y^-)\},\\
U_i(y)&=F_{1,i}^{\mathrm{all}}(y)+0.25\,F_{1,i}^{E}(y),
\end{aligned}
\end{equation}
where $F_{1,i}^{\mathrm{all}}(y)$ and $F_{1,i}^{E}(y)$ are the Combined and
Edge F1 of candidate $y$ against the reference of item $i$
(Equation~\ref{eq:eai_f1}).
Here $m=0.10$ multiplies the quality gap; it is neither subtracted from the
gap nor added as a separate constant. Thus the smallest pair margin is
$0.10\times0.01=0.001$. Minimising the loss favours a larger preferred-minus-rejected
score difference. Gradients update both candidate scores; utilities and gaps
remain fixed. This best-versus-lower-quality pairwise objective does not model
a listwise permutation probability.
The SFT term preserves reference supervision while the ranking term
favours higher-quality complete goals. When no lower-utility candidate exists, the ranking term is zero.
Parameter settings and the training-only search protocol are reported below.

\paragraph{Worked example}
Figure~\ref{fig:grace_framework} follows diagnostic record \texttt{783\_1}
through both stages. Its pool contains three unique candidates, which differ
only in their edge goals. Candidate~1, the greedy output, omits the
\texttt{dish\_soap} relation (Combined F1 0.889, Edge F1 0.800); Candidate~2
also omits \texttt{cup} (0.750, 0.500); Candidate~3 matches the reference
(1.000, 1.000). Before CORE, the sequence scores are $-0.0173$, $-0.0221$,
and $-0.0143$, so likelihood selection already returns the complete goal
that greedy decoding misses. After CORE training the scores are $-0.0173$,
$-0.0236$, and $-0.0139$: the weakest candidate falls, the complete one
rises, and the selection is unchanged. As average per-token probabilities
$\exp(s_\theta)$, the form shown in the figure, these scores are 98.29\%,
97.81\%, and 98.58\% before and 98.28\%, 97.67\%, and 98.62\% after. CORE reranks the same archived
candidates; it does not regenerate them. One record illustrates the mechanism
only; Table~\ref{tab:main_results} gives the 68-record scores, where the
Edge component is lowest throughout. No reference is read at inference; F1 is computed
afterwards. During training the same construction is applied to training
records only: candidates are model-generated, scored against training
references, and paired when their quality differs. The ranking term
raises the preferred candidate's relative sequence score while the SFT term
retains the reference target. CORE thus ranks complete candidates and selects
one, rather than classifying the input into a fixed set of task labels.

\section{Experiments}
\label{sec:experimental_design}
\subsection{Dataset and Split}

The study uses the 342 public Goal Interpretation records from EAI
\cite{li2024embodied}, whose household scenes derive from VirtualHome
\cite{puig2018virtualhome}. Each record contains an instruction, the relevant
scene and schema, and a sparse reference goal with Node states, directed Edge
relations, and terminal Actions. The target represents success conditions, not
an action sequence. Prompts and reference semantics are retained unchanged;
identifier, parsing, duplication, and split-conservation checks verify the
materialised corpus.

Task families are grouped before splitting. Twenty-one families (274 records)
form the adaptation set; five families (68 records) form the diagnostic set.
Table~\ref{tab:dataset_split} reports their composition.

\begin{table}[t]
\centering
\caption{The 342-item EAI goal interpretation corpus and task-grouped 274/68 split.}
\label{tab:dataset_split}
\footnotesize
\begin{tabular}{@{}llr@{}}
\toprule
Split & Task family & Items\\
\midrule
Public corpus & All 26 task families & 342\\
Training & 21 task families & 274\\
Diagnostic & Change TV channel & 3\\
Diagnostic & Drink & 23\\
Diagnostic & Listen to music & 19\\
Diagnostic & Wash dishes by hand & 10\\
Diagnostic & Wash dishes with dishwasher & 13\\
\midrule
Diagnostic total & 5 task families & 68\\
\bottomrule
\end{tabular}
\end{table}

There is no identifier, task-name, or exact-target overlap. Because the 68
records were repeatedly inspected, they are treated as a task-disjoint
diagnostic rather than an untouched test set. Large-model reproduction uses all
342 records, whereas adaptation and learned ranking use the 274/68 split.

\subsection{Parameter Settings}

SFT adapts Qwen2.5--7B-Instruct \cite{qwen2024report} with LoRA for three epochs, learning rate
$2\times10^{-4}$, and seed 42; only assistant tokens contribute to the loss.
Candidate texts and reference-derived training utilities are fixed before
preference training; model likelihood scores are recomputed and updated. CORE
continues SFT for one epoch with learning rate $10^{-5}$, gradient accumulation
eight, $\alpha=\gamma=1.000$, and $m=0.100$, producing 73 comparisons and 35
updates. DPO uses 41 aligned pairs, one epoch, learning rate $5\times10^{-6}$,
$\beta=0.100$, and the same diagnostic exclusion.

The archived search configuration specifies task-grouped five-fold search
within the 274 adaptation records, with $\gamma\in\{0.03,0.10,0.30,1.00\}$
and $m\in\{0,0.05,0.10,0.20\}$ in a staged search. The final run verifies
$\gamma=1$ and $m=0.10$; detailed fold scores are not available in the retained
archive, so we do not claim a measured sensitivity curve. The anchor weight
$\alpha=1$, Edge utility weight $0.25$, and gap floor $0.01$ are fixed design
choices, not independently validated optima. The Edge term prioritises
relational errors; the floor supplies a minimum positive quality gap.

The fixed diagnostic bank was generated by an earlier ranking continuation
(rank weight $0.10$), not the pure SFT checkpoint. It uses one greedy output
and four samples at each of two temperature and top-$p$ settings, $(0.65,0.90)$ and $(0.90,0.95)$, with a
2048-token limit and exact-string deduplication. CORE reranks this same
archived bank. This controls candidate availability for that reranking
contrast, but does not isolate the source of the bank or end-to-end training
effects against greedy SFT. Our inference ablation keeps the earlier checkpoint
fixed. A shared bank from the exact SFT parent would be needed for a matched
selector study; separate banks per selector would instead measure complete
generation-and-selection pipelines.

\subsection{Evaluation Metrics}

For component $k\in\{N,E,A\}$ (Node, Edge, Action), the released EAI
evaluator accumulates true positives $TP_k$, false positives $FP_k$, and false
negatives $FN_k$:
\begin{equation}
\label{eq:eai_f1}
\begin{aligned}
F_{1}^{k}&=\frac{2TP_k}{2TP_k+FP_k+FN_k},\\
F_{1}^{\mathrm{all}}&=\frac{2\sum_kTP_k}
 {2\sum_kTP_k+\sum_kFP_k+\sum_kFN_k}.
\end{aligned}
\end{equation}
Thus Combined F1, $F_{1}^{\mathrm{all}}$, is a micro-aggregation of counts, not the arithmetic mean of
the three component F1 values.

Predictions must parse and contain 68 unique aligned identifiers. The official
EAI evaluator reports Node, Edge, Action, and micro-aggregated Combined F1 as
defined in Equation~\ref{eq:eai_f1}. Candidate-oracle quality and
selection changes diagnose why the final F1 changes. All headline open-model results use the unmodified original evaluator;
the separately labelled large-model reproduction uses its no-closure protocol.

Split, model revision, predictions, candidate banks, scores, and preference
pairs are hash-recorded. Diagnostic gold is used only after
inference for official scoring, oracle analysis, and error classification; it
is unavailable to deployed selection. Scores with different
evaluation scopes are not treated as controlled comparisons. The study reports
one optimisation/sampling seed (42), distinct from the fixed task-family split,
and therefore makes no confidence-interval or significance claim;
failed and gate-stopped experiments remain part of the audit trail.

\subsection{Comparison Methods}
We report DeepSeek-V3 Chat \cite{deepseek2024v3}, a directly prompted LLM, and
DeepSeek-Reasoner, which reasons step by step before answering, as
full-corpus, no-closure reproductions. Archived raw Qwen2.5--7B and 72B
inference results provide an additional model-family reference under the
original EAI protocol. These models are evaluated without additional training;
their 342-item scope differs from the adapted-model diagnostic.

The 68-item configuration comparison contains greedy SFT, length-normalised
likelihood ranking with the earlier ranking-trained checkpoint, candidate-level DPO, and our
SFT+CORE approach. All use the same diagnostic identifiers and original EAI
evaluator. BRIO and SLiC-HF motivate sequence-level ranking but are not
additional reproduced baselines \cite{liu-etal-2022-brio,zhao2023slichf}.

\paragraph{KG-based comparators}
We include two distinct knowledge-based routes. The unadapted Qwen2.5--7B
KG-prompt arm prepends scene-relevant schema hints (object states, relation
constraints and implications) to the task prompt, without SFT or CORE.
We score its archived raw response, without the older pipeline's deductive
closure or refinement, to isolate this prompt-only configuration.
The Euclidean and hyperbolic KG arms instead combine standard SFT with
projected KG prefix representations, without CORE ranking. Their archived
runs use the same 274/68 task split, three epochs, learning rate
$2\times10^{-4}$ and seed 42. They are KG-enhanced language models, not
standalone symbolic KG solvers. We re-evaluate all archived outputs on the
same 68 identifiers using the current original-EAI resources; re-evaluation
reproduces the SFT baseline. These are configuration comparisons with differing
training and inference procedures, not isolated causal estimates of KG utility.

\paragraph{Candidate-level DPO}

The validated preference-learning arm uses model-generated,
candidate-level pairs but optimises Direct Preference Optimisation (DPO)
\cite{rafailov2023direct}. It also trains on positive--negative pairs
$(y_i^{+},y_i^{-})$, but builds them differently from CORE: the frozen parent
output is the negative $y_i^{-}$, and a complete candidate is taken as the
positive $y_i^{+}$ only when it has higher canonical Combined F1 and its Edge
F1 is no more than $0.050$ below that of $y_i^{-}$. This conservative rule yields 41 pairs from the 274 training records.
DPO increases the preferred response's log-probability relative to the rejected
response and a frozen reference policy \cite{rafailov2023direct}; scores cover
all assistant tokens. Settings are listed above. The fixed diagnostic is
excluded from pair construction and model selection. Because DPO and CORE use
41 and 73 comparisons respectively and different learning rates, their
comparison does not isolate objective choice alone.

\paragraph{Candidate oracle}
After inference, a candidate oracle uses the reference-based quality $Q_i(y)$
of each candidate (Combined F1, with ties broken by Edge F1) and selects
\begin{equation}
\label{eq:candidate_oracle}
y_i^{\mathrm{oracle}}=\arg\max_{y\in\mathcal C_i}Q_i(y).
\end{equation}
It is not deployable;
it measures candidate headroom and separates generation from selection
failures.

\section{Results and Discussion}
\label{sec:results}
\subsection{Large-Model and Adapted-Qwen Results}

The reasoning-specialised model improves the unassisted score from 0.318 to
0.434. Additional reasoning therefore helps, but does not by itself solve
EAI's sparse symbolic representation problem. Since model scale, training data,
and reasoning procedure are confounded, this is not a causal scaling claim.

Table~\ref{tab:context_results} reports full-corpus model results separately
from the adapted-model comparison. Raw Qwen2.5--72B obtains 0.293 Combined F1
in the archived original-EAI evaluation, providing a second model family
alongside DeepSeek. Different evaluation protocols prevent a controlled
cross-family ranking. Likewise, raw full-corpus and adapted diagnostic scores
do not quantify a matched SFT effect.

\begin{table}[t]
\centering
\caption{Contextual model results on 342 items. NC: no-closure protocol;
Orig.: original EAI. Rows with different protocols are not controlled comparisons.}
\label{tab:context_results}
\scriptsize
\setlength{\tabcolsep}{2.8pt}
\begin{tabular}{@{}llrrrr@{}}
\toprule
Model & Protocol & Comb. & Node & Edge & Action\\
\midrule
DeepSeek-V3 Chat & NC & 0.318 & 0.394 & 0.318 & 0.233\\
DeepSeek-Reasoner & NC & 0.434 & 0.429 & 0.526 & 0.345\\
\midrule
Qwen2.5--7B & Orig. & 0.243 & 0.372 & 0.238 & 0.141\\
Qwen2.5--72B & Orig. & 0.293 & 0.387 & 0.308 & 0.198\\
\bottomrule
\end{tabular}
\end{table}

Table~\ref{tab:main_results} compares adapted models on the same 68-item
diagnostic. CORE improves Combined F1 by 0.020 and Edge F1 by 0.035 over
greedy SFT. Its 0.001 Combined advantage over DPO is descriptive; one seed
and a repeatedly inspected diagnostic do not support a significance claim.
DPO remains strongest on Action F1.

\begin{table}[t]
\centering
\caption{Configuration comparison on the same 68-item diagnostic, original EAI F1.
Training conditions differ; this is not a loss-only ablation.
Bold marks the best value in each column, including ties.}
\label{tab:main_results}
\scriptsize
\setlength{\tabcolsep}{3pt}
\begin{tabular}{@{}lrrrr@{}}
\toprule
Method & Combined & Node & Edge & Action\\
\midrule
SFT baseline (greedy) & 0.716 & 0.880 & 0.606 & 0.628\\
Earlier ranker + selection & 0.726 & \textbf{0.887} & 0.615 & 0.653\\
Candidate-level DPO & 0.735 & \textbf{0.887} & 0.613 & \textbf{0.681}\\
\midrule
\textbf{SFT + CORE (ours)} & \textbf{0.736} & \textbf{0.887} & \textbf{0.642} & 0.653\\
\bottomrule
\end{tabular}
\end{table}

\subsection{Comparison with KG-Based Methods}
Table~\ref{tab:kg_comparison} compares the archived KG configurations under
one diagnostic/evaluator scope. KG prompting scores below its unadapted
baseline, while both KG-prefix SFT arms score below plain SFT. CORE has the
highest Combined and Edge F1 among these configurations. This result concerns
the implemented knowledge interfaces and does not establish that KGs are
ineffective in general. In particular, unadapted prompting and trained ranking
use different supervision budgets.

\begin{table}[t]
\centering
\caption{KG-based configuration comparison on the same 68 items, original EAI
F1. Archived raw outputs re-evaluated; training and inference procedures differ.}
\label{tab:kg_comparison}
\scriptsize
\setlength{\tabcolsep}{3pt}
\begin{tabular}{@{}lrrrr@{}}
\toprule
Configuration & Comb. & Node & Edge & Action\\
\midrule
Qwen7B, unadapted & 0.258 & 0.328 & 0.173 & 0.253\\
Qwen7B + KG prompt & 0.199 & 0.237 & 0.143 & 0.217\\
\midrule
SFT, no KG & 0.716 & 0.880 & 0.606 & 0.628\\
SFT + Euclidean KG & 0.616 & 0.885 & 0.338 & 0.589\\
SFT + hyperbolic KG & 0.627 & 0.880 & 0.468 & 0.430\\
\midrule
SFT + CORE (ours) & \textbf{0.736} & \textbf{0.887} & \textbf{0.642} & \textbf{0.653}\\
\bottomrule
\end{tabular}
\end{table}

\subsection{Candidate Selection Analysis}

For the earlier ranking-trained checkpoint, length-normalised sequence ranking changes 12/68 greedy selections and raises
all four components. Candidate sets contain 1--9 unique outputs (mean 3.353).
An Edge-level analysis partitions the items into 38 exact selected sets, 11 generation
failures with no improved candidate, and 19 recoverable selection failures. A
better Edge candidate lies within the sequence-score top three for 16/19, but
the selector chooses an improvement in only 6/19. The 0.097 Combined gap
between sequence ranking and the gold-selected oracle locates substantial
unrealised information in the candidate bank. The non-deployable oracle reaches
0.823 Combined (0.957 Node, 0.752 Edge, 0.692 Action); it is an analysis upper
bound, not an experimental method.

The analysis suggests different interventions: improve candidate generation when
no better goal is available, and improve ranking when it is already present.

\subsection{Ablation: Candidate-Output Ranking}

\begin{figure}[t]
\centering
\includegraphics[width=\columnwidth]{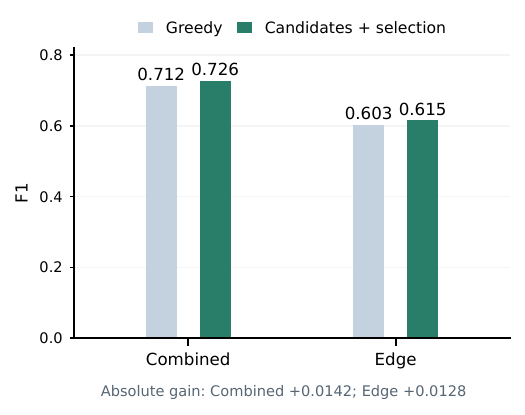}
\Description{Bar chart with two groups, Combined F1 and Edge F1, each comparing
greedy decoding with candidate generation plus likelihood selection for the
same earlier ranking-trained checkpoint. Combined F1 rises from 0.712 to 0.726
and Edge F1 from 0.603 to 0.615.}
\caption{Matched inference ablation on the 68-item diagnostic.
Both variants use the same earlier ranking-trained checkpoint (weight 0.10).
The F1 axis is linear and starts at zero. Bar labels report the actual
F1 scores; absolute gains are 0.014247 Combined and 0.012821 Edge F1.}
\label{fig:ranking_ablation}
\end{figure}

Figure~\ref{fig:ranking_ablation} isolates the inference procedure using the
same earlier ranking-trained checkpoint. Greedy decoding scores 0.712121
Combined and 0.602564 Edge F1; candidate generation plus likelihood selection
scores 0.726368 and 0.615385, gains of 0.014247 and 0.012821 respectively.
This jointly tests sampling and selection, not ranking-loss training. The
earlier checkpoint's greedy score is below pure SFT (0.716113 Combined), so
extra ranking training alone cannot be credited with the inference gain.
A matched extra-SFT control and independent anchor/margin ablations remain
unperformed; the final CORE comparison does not substitute for these controls.

CORE combines an SFT anchor with candidate-level contrastive ranking. The
final run uses rank weight 1.000 and margin 0.100;
the final 274-record run forms 73 candidate comparisons. CORE reaches 0.736
Combined and 0.642 Edge F1, improving every SFT component and exceeding DPO by
0.001 Combined and 0.028 Edge. DPO remains stronger on Action (0.681 versus
0.653). The comparison evaluates the implemented preference-learning configurations;
their different pair counts and learning rates do not isolate the loss alone.

\section{Conclusion}
\label{sec:conclusion}
We presented a two-stage generate-then-rank approach to goal interpretation
for embodied agents: supervised fine-tuning for goal generation, followed by
learning to rank candidate outputs with CORE.
On the fixed diagnostic, CORE improves Combined and Edge F1 over greedy SFT
and the implemented DPO comparator, while DPO retains higher Action F1.
The candidate analysis identifies useful goals that generation already supplies
but selection misses.

The evidence is limited to one adaptation seed and a repeatedly inspected
68-item diagnostic covering five task families. CORE and DPO also differ in
pair construction and optimisation settings, and CORE inference
reuses a bank generated by an earlier ranking-trained checkpoint. These
comparisons do not establish an isolated causal effect of the CORE loss.
The anchor and Edge utility weights lack sensitivity ablations, and
serialisation robustness is untested.

The immediate experimental priority is a matched extra-SFT versus CORE
continuation from the same parent, holding data, update budget, learning rate,
and candidate bank fixed, followed by anchor/Edge-weight ablations. Multiple
training seeds can retain the exact task-disjoint 274/68 split; no diagnostic
items need enter training. Such runs measure optimisation variability, while
fresh held-out families are needed to assess generalisation beyond this
repeatedly inspected diagnostic. Listwise ranking and matched constrained
decoding remain untested alternatives.
Downstream planning success and semantic repair require separate evaluation:
symbolic F1 alone cannot establish executability.

\begin{credits}
\subsubsection{\discintname}
The authors have no competing interests to declare that are relevant to the
content of this article.
\end{credits}
%
%
%
\bibliographystyle{splncs04}
\bibliography{references}
\end{document}